\documentclass{article}

\usepackage{microtype}
\usepackage{graphicx}
\usepackage{subfigure}
\usepackage{booktabs} 

\usepackage{hyperref}

\usepackage[accepted]{icml2020}
\usepackage{bm}

\newcommand{\eg}{\emph{e.g.,}\ }

\newcommand{\wrt}{\emph{w.r.t.}\ }

\icmltitlerunning{KGNN: Distributed Framework for Graph Neural Knowledge Representation}

\begin{document}

\twocolumn[
\icmltitle{KGNN: Distributed Framework for Graph Neural Knowledge Representation}





\begin{icmlauthorlist}
\icmlauthor{Binbin Hu}{ant}
\icmlauthor{Zhiyang Hu}{ant}
\icmlauthor{Zhiqiang Zhang}{ant}
\icmlauthor{Jun Zhou}{ant}
\icmlauthor{Chuan Shi}{bupt}

\end{icmlauthorlist}

\icmlaffiliation{ant}{Ant Financial Services Group, Hangzhou, China}
\icmlaffiliation{bupt}{Beijing University of Posts and Telecommunications, Beijing, China}
\icmlcorrespondingauthor{Binbin Hu}{bin.hbb@antfin.com}

\icmlkeywords{Machine Learning, ICML}

\vskip 0.3in
]



\printAffiliationsAndNotice{}  

\begin{abstract}
Knowledge representation learning
has been commonly adopted to incorporate knowledge graph (KG) into various online services. Although existing knowledge representation learning methods have achieved considerable performance improvement, they ignore high-order structure and abundant attribute information, resulting unsatisfactory performance on semantics-rich KGs. Moreover, they fail to make prediction in an inductive manner and cannot scale to large industrial graphs. To address these issues, we develop a novel framework called KGNN to take full advantage of knowledge data for representation learning in the distributed learning system. KGNN is equipped with GNN based encoder and knowledge aware decoder, which aim to jointly explore high-order structure and attribute information together in a fine-grained fashion and preserve the relation patterns in KGs, respectively. Extensive experiments on three datasets for link prediction and triplet classification task demonstrate the effectiveness and scalability of KGNN framework.
\end{abstract}

\section{Introduction}

\begin{figure}
	\begin{center}
		\includegraphics[width=0.85\linewidth]{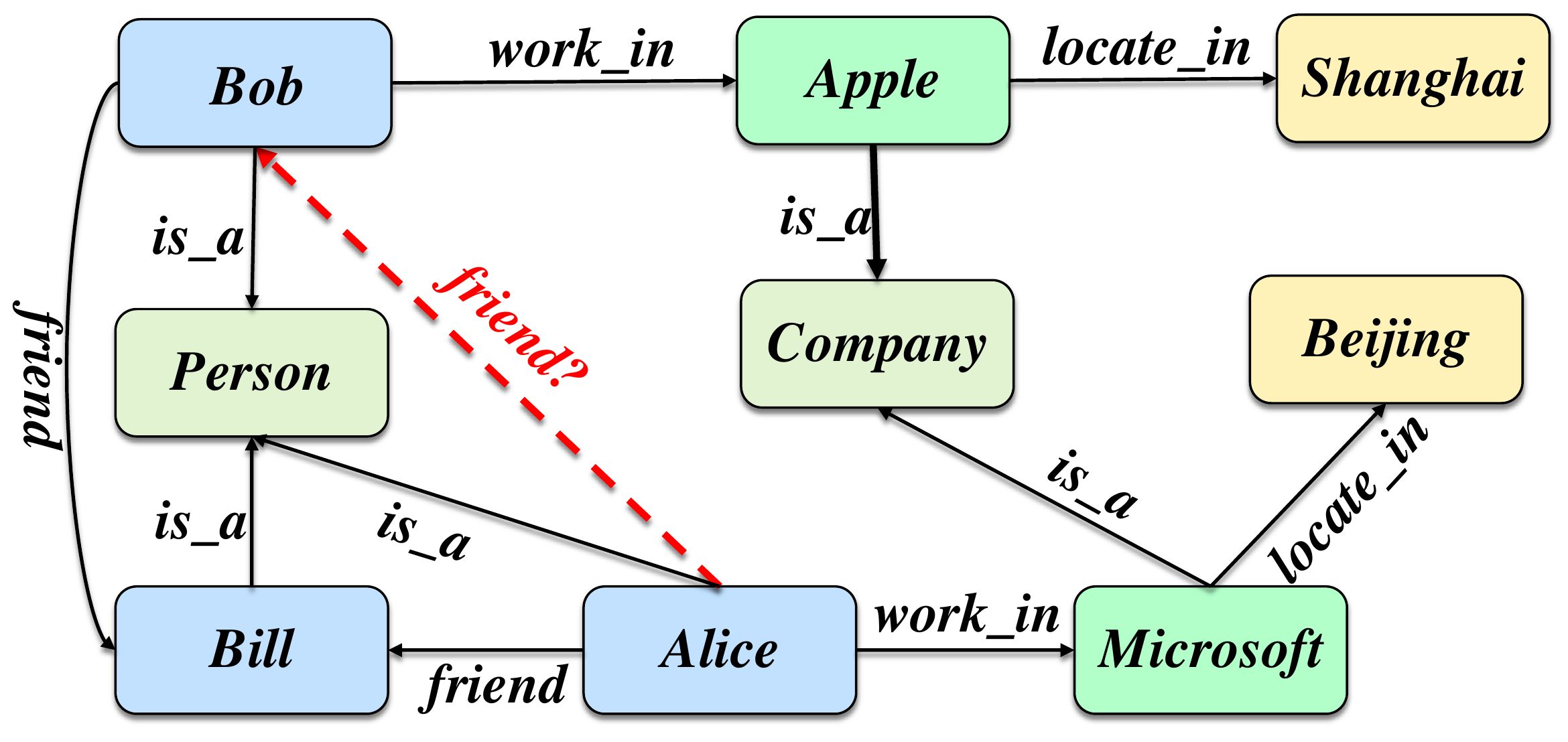}
	\end{center}
	\caption{The example of knowledge graph.}
	\label{fig:kg}
\end{figure}

Knowledge graph (KG) represents the heterogeneous structure of entities and their rich relations in triplets of the form $\langle head\ entity, relation, tail\ entity \rangle$. For example in Fig.~\ref{fig:kg}, a triplet $\langle Bob, work\_in, Apple \rangle$ is denoted as a relation $work\_in$ connecting two entities: $Bob$ and $Apple$. Due to abundant structured information, KG has attracted much attention in many research areas, ranging from information retrieval~\cite{dietz2018utilizing}, question answering~\cite{huang2019knowledge} to recommender system~\cite{cao2019unifying}.

To flexibly incorporate such knowledge, knowledge representation learning~\cite{wang2017knowledge} has emerged as a promising direction for knowledge completion~\cite{lacroix2018canonical}, alignment~\cite{wang2018cross} and reasoning~\cite{trivedi2017know}, which aims to project both entities and relations into a low-dimensional space whilst preserving certain information of the original graph. These methods can be broadly classified as  \emph{translational distance models}~\cite{bordes2013translating,wang2014knowledge,lin2015learning,sun2019rotate} and \emph{semantic matching models}~\cite{nickel2011three,jenatton2012latent,yang2014embedding,trouillon2016complex,dettmers2018convolutional}, which exploit distance-based and similarity-based scoring function for knowledge representation learning, respectively. 

Although these methods have yield considerable performance improvements to some extent, they still suffer from several limitations. First, they process each triple independently and abundant attributes in nodes and edges are commonly ignored, resulting in unsatisfactory performance on semantics-rich KGs. Second, they are inherently transductive models, which cannot make prediction for entities unseen in the training set. Third, these methods cannot scale to industral-scale graphs with hundreds of millions of entities and relations.

To address these issues, in this paper, we aim to build a scalable and distributed knowledge graph representation framework to flexibly distill rich knowledge for downstream applications. Intuitively, the framework is expected to satisfy the following three key properties: 
(1) \textbf{Semantics-rich}: High order structure and attribute information have been already proved effective for preserving properties of original graphs in previous works~\cite{xu2018representation,abu2019mixhop,luan2019break}
. Hence, we aim to incorporate such information into knowledge graph representation to comprehensively capture rich semantics in KGs. 
(2) \textbf{Inductive}: Current KGs are usually far from complete and thus the new entities will appear everyday in the real-world setting, which prompts the proposal to make prediction for entities unseen in the training set dynamically.
(3) \textbf{Scalable}: Since KGs in the real-world industrial scenarios are extremely large-scale, a scalable knowledge graph representation framework implemented on distributed learning system is in urgent demand.

To integrate above main idea together, we propose \textbf{KGNN}, a distributed framework for graph neural knowledge representation with graph neural network (GNN) based encoder and knowledge aware decoder. With the help recently emerging GNN, KGNN is potential to jointly capture attribute information and high order structure in an inductive, end-to-end framework. Obviously, it is a flexible framework to equip arbitrary GNN based encoder, and in this paper, an attention based GNN is introduce to locate the important and relevant relations or structures for fine-grained semantics.  In order to perform model training and inference effectively for real-world KGs, KGNN is implemented on the distributed learning system and the implementation details are uncovered. We make extensive experiments on three real-world datasets on link prediction and triplet classification task,  which demonstrates the effectiveness and scalability of the proposed KGNN framework.

\section{Background}
In this section, we give a brief overview of knowledge representation learning and graph neural networks.

\textbf{Knowledge representation learning.}
A knowledge graph is denoted by $\mathcal{G} = \{ \mathcal{E}, \mathcal{R} \}$, consisting of the entity set $\mathcal{E}$ and the relation set $\mathcal{R}$. A triplet $\langle h, r, t \rangle$ is defined as an relation $r$ between entities $h$ and $t$ on $\mathcal{G}$, where $h, r \in \mathcal{G}$. 
Learning distributional representations of KGs provides an effective and efficient way for applying structural knowledge in various applications. 
Hence, a scoring function $s(e_h, e_r, e_t)$ is defined as the likelihood of triple  $\langle h, r, t \rangle$ being a valid triple, where $e_h, e_r, e_t$ represent the embeddings of $h, r, t$, respectively. A series of scoring functions~\cite{wang2017knowledge} are proposed to preserve different relation patterns of KGs, and here, we introduce the \emph{TransH} based scoring function~\cite{wang2014knowledge}, which learns different representations for an entity conditioned on different relations. 
\begin{equation}
\label{equ_score}
    s(e_h, e_r, e_t) = ||e^{\perp}_h + e_r - e^{\perp}_t||.
\end{equation}
Here, we have $e^{\perp}_h = e_h - w_r^Te_hw_r$ and $e^{\perp}_t = e_t - w_r^Te_tw_r$, in order to project entity embeddings into relation heperplanes, which allows entities playing different roles under different relations.

\textbf{Graph neural network.} Graph neural network (GNN) makes use of  the structure of the graph and  attributes on nodes for representation learning~\cite{hamilton2017inductive, velivckovic2017graph, kipf2016semi}. Specifically, GNN recursively update an node's representation by aggregating information from its neighbors. Subsequently, the final representations of the nodes after $k$ updating capture the structural information as well as the node attributes within $k$-hop neighbors. Formally, we can calculate the $k + 1$-th representation for node $v$ with aggregation and updating function as follows,
\begin{equation}
\label{equ_rep}
    e_v^{k + 1} = f^{(U)} (e^{k}_v, f^{(A)}(\{e^{k}_{v'}, v' \in \mathcal{N}_v\}; \Theta^{(A)})  ; \Theta^{(U)} ),
\end{equation}
where  $f^{(A)}$ and $ f^{(U)}$ denotes the aggregation and updating function parameterized by $ \Theta^{(A)}$ and $\Theta^{(U)}$, respectively, and $\mathcal{N}_v$ is the neighbor set of node $v$.

\section{Methodology}

In this section, we present the distributed framework for graph neural knowledge representation, called \textbf{KGNN}.

\begin{figure}
	\begin{center}
		\includegraphics[width=1.0\linewidth]{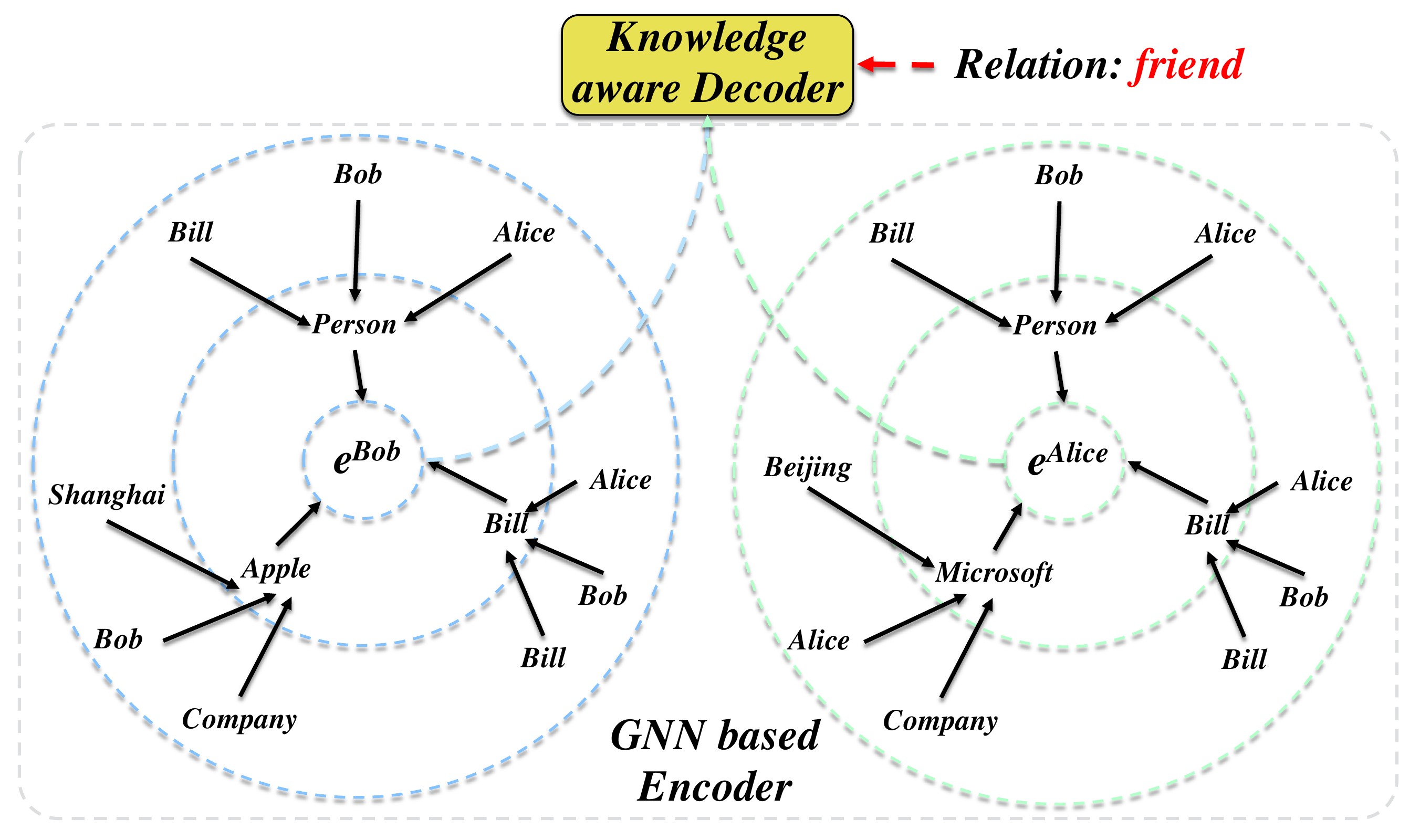}
	\end{center}
	\caption{Overview of KGNN model.}
	\label{fig_kgnn}
\end{figure}

\subsection{KGNN Model}
\label{sec_kgnn}

In this section, we introduce the model part of KGNN to comprehensively distill knowledge graph for representation learning in an inductive manner. We present the architecture of our proposed KGNN in Fig.~\ref{fig_kgnn}, which intuitively consists of two modules: (1) GNN based encoder and knowledge aware decoder, which flexibly utilizes the local structure information and recursively propagates the embeddings over KGs for expressive representations and (2) knowledge aware decoder, which aims to preserve the relation patterns in KGs through various types of score functions. 

\textbf{GNN based Encoder.}
Different from one-hot representation in previous works, we propose to adopt graph neural network to encode structural knowledge and attributes into entities' representations. For fine-grained modeling , we introduce an attention based GNN to weighs various underlying preference for each relation. Following the above updating principle of entity representations in Eq.~\ref{equ_rep}, we firstly formulate the aggregation function $f^{(A)}(\cdot)$ as follows:
\begin{equation}
    \label{equ_agg}
    f^{(A)}(\{e^{k}_{t}, t \in \mathcal{N}^k_h\}) = \sum_{r, t \in \mathcal{N}_h}\alpha(h, r, t)e^{k}_t.
\end{equation}
Here, $\alpha(h, r, t)$ is the attention value for the triple $\langle h, r, t \rangle$, which is implemented as a neural network. And $\mathcal{N}^k_h = \{ (r, t) | (h, r, t) \in \mathcal{G}\}$ is the $k$-hop neighbor set for entity $h$. 

\begin{figure}
	\begin{center}
		\includegraphics[width=0.95\linewidth]{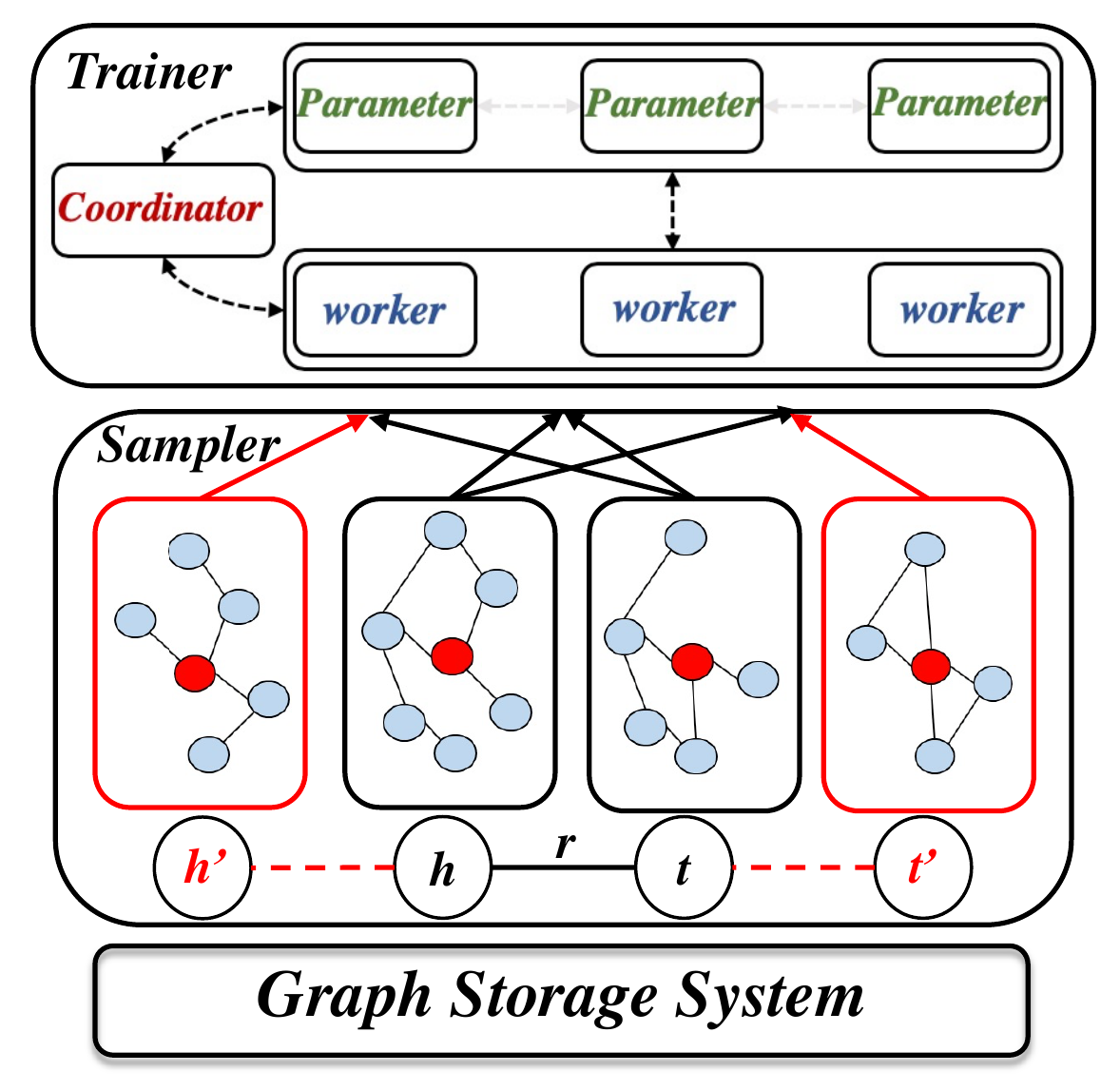}
	\end{center}
	\caption{Overview of distributed KGNN framework.}
	\label{fig_kgnn_dis}
\end{figure}
Inspired by the idea jumping knowledge network~\cite{xu2018representation}, we adopt an adaptive depth function to flexibly multiple hops of neighbors  for better structure-aware representation. Here, an LSTM is applied to implement $f^{(U)}(\cdot)$ for representation updating. Therefore, we can obtain the $k+1$-th representation for entity $h$ as follows:
\begin{equation}
    e^{k+1}_h = {LSTM}(e^k_h, a^k_h),
\end{equation}
where $a^k_h$ denotes the aggregated information for entity $h$, calculated by Eq.~\ref{equ_agg}.

\textbf{Knowledge aware decoder.}
The key of link prediction in KGs is to infer the relation patterns {\eg symmetry, inversion and composition} with observed triplets~\cite{sun2019rotate}. In order to adaptively preserve different relation patterns on various KGs, KGNN adopts knowledge aware score function as the decoder. Take the \emph{TransH} as an example, we represent the score function for a triple $\langle h, r, t\rangle$ after $K$-hop updating as $s(e^K_h, e_r, e^K_t)$. Then, we train KGNN in an end-to-end fashion via the margin based objective with negative sampling:
\begin{equation}
    \mathcal{L} = \sum_{\langle h, r, t \rangle \in \mathcal{G}, \langle h', r, t' \rangle \in \mathcal{G'}}[s(e^K_h, e_r, e^K_t) +\lambda -s(e^K_{h'}, e_r, e^K_{t'})]_+,
\end{equation}
where $[\cdot] = max(0, \cdot)$, and $\mathcal{G}'$ is the set of incorrect triplets constructed by 
randomly replacing head entity or tail entity in a valid triplet.

\subsection{Distributed Implementation}
We now zoom into the distribution implementation of KGNN, which provide a complete solution for large-scale  knowledge graph representation. As shown in Fig.~\ref{fig_kgnn_dis}, the distributed KGNN is comprised of three parts: 
\begin{itemize}
    \item \textbf{Graph storage system}. It stores the whole knowledge graph as well as corresponding attributes information on nodes under the distributed architecture. With the help of the effective data compression technology, it is capable of serving large-scale industrial graphs. 
    
    \item \textbf{Sampler}. It mainly provides negative sampler and sub-graph sampler for knowledge representation. In particular, the negative sampler randomly replaces head entity or tail entity in a batch of valid triplets for corresponding corrupted triplets. And then, sub-graph sampler will randomly collect $k$-hop neighbors set for each entity in batch. It is worth noting that we feed the sub-graph into KGNN instead of the full graph, which helps reduce the time and memory cost.
    
    \item \textbf{Trainer}. It consisting of several workers and parameter servers, controlled by the coordinator. For effective parameter updating, each work pulls parameters from a parameter server and update them independently during training. In a specific worker, KGNN naturally follows such a work flow: (1) Pre-process the sub-graph and parse the model config. (2) Produce embeddings for entities and relations based on sub-graph with our encoder and decoder introduced in Sec.~\ref{sec_kgnn}. (3) Optimize a certain loss to guide the learning process.
 
\end{itemize}




\section{Experiments}

In this section, we evaluate the effectiveness of KGNN for link prediction and triplet classification task.


\begin{table}[]
    \centering
    \caption{Statistics of data sets.}
    \begin{tabular}{c|c|c|c|c}
        \hline
        {Dataset} & {\# Ent.} & {\# Rel.} & {\# Trip.}  & {\# Attr.}\\
        \hline 
        {WN18} & {40, 943} & {18} & {151, 442} & {N.A.} \\
        \hline
        {FB15K} & {14, 951} & {1, 345} & {592, 213} & {N.A.}\\
        \hline
         {Alipay} & {$2.6 \times 10^5$} & {6} & {$1.28 \times 10^6$} & {504}\\
        \hline
    \end{tabular}
    \label{tab_data}
\end{table}

\begin{table}[h]
	\centering
	\caption{Evaluation results on link prediction. (\%)}
	\begin{tabular}{c|c|c|c|c|c|c}
		\hline
		{Methods} & \multicolumn{3}{|c|}{WN18 (HR@k)} &  \multicolumn{3}{|c}{FB15K  (HR@k)} \\
		\hline
	    {} & {1} & {3} & {10} & {1} & {3} & {10} \\
	    \hline
	    {TransE} & {70.8} & {89.6} & {94.7} & {64.4} & {84.0}& {95.9} \\
		
		{TransR} & {65.5} & {83.7} & {92.7} & {63.6} & {82.3}& {95.2} \\
		
		{TransH} & {72.3} & {90.6} & {94.9} & {64.5} & {84.1}& {95.9} \\
		
		{DistMult} & {69.3} & {89.9} & {94.6} & {65.2} & {84.6}& {96.5} \\
		\hline
		{KGNN} & {\textbf{78.9}} & \textbf{96.9} & {\textbf{98.8}} & {\textbf{67.4}} & {\textbf{86.4}}& {\textbf{96.8}} \\
		\hline
	\end{tabular}
	\label{tab_link_pred}
\end{table}

\textbf{Datasets and evaluation metrics.}
We evaluate our proposed framework on three datasets~\cite{lin2015learning}, namely WN18, FB15K and industrial AliPay dataset. The detailed  descriptions of the three datasets are summarized in Tab.~\ref{tab_data}. We perform link prediction on WN18 and FB15K, while apply triple classification on WN18, FB15K and AliPay dataset. Following the same setting in \cite{bordes2011learning,bordes2013translating}, we adopt hit ratio at rank $k$ (HR@k) and area under ROC curve (AUC) to evaluate the model performance of link prediction and triplet classification,  respectively.

\begin{table}[h]
    \centering
    \caption{AUC comparison results on triplet classification. (\%)}
    \begin{tabular}{c|c|c|c}
         \hline
		{Methods} & {WN18} & {FB15K} & {Alipay} \\
         \hline
		{TransE} & {91.7} & {97.5} & {61.0} \\
	
		{TransR} & {78.6} & {95.8} & {74.9} \\
		 
		{TransH} & {91.7} & {97.4} & {72.6} \\
		 
		{DistMult} & {93.5} & {97.9} & {65.3} \\
		 \hline
		 \hline
		{KGNN} & {\textbf{94.1}} & {\textbf{99.0}} & {\textbf{84.9}} \\
		\hline
    \end{tabular}
    
    \label{tab_classification}
\end{table}

\textbf{Compared methods and parameter settings.}
We consider for representative knowledge representation learning methods for performance comparison, namely TransE~\cite{bordes2013translating}, TransR~\cite{lin2015learning}, TransH~\cite{wang2014knowledge} and DistMult~\cite{yang2014embedding}. For fair comparison, we also select one of them as the decoder of KGNN framework. We adopt Adam with learning rate = 0.001 to optimize all models and set the batch size = 256. Moreover, the margin is selected among \{1, 2, 5\} and the embedding size is searched among \{64, 128, 256\}.

\textbf{Performance Comparison.}
We report the comparison results of the proposed KGNN and baselines on link prediction and triplet classification in Tab.~\ref{tab_link_pred} and Tab.~\ref{tab_classification}, respectively. We observe that KGNN consistently outperform on three datasets for both tasks, indicating that KGNN is potential to capture high-order structural information for more expressive knowledge representations. It is worthwhile to note that KGNN achieve significant performance improvement over baselines on Alipay datasets. The results may correlated with the characteristics of this dataset: (1) There are 504 attributes on entities, which are ignored by these baselines. (2) The test set contains a part of unseen entities, while these baselines fail to produce proper representations for them. As a comparison, the performance of KGNN demonstrates that KGNN is capable of jointly exploring structure and attribute information together over KGs in an inductive manner.

\begin{figure}[h]
	\centering
	\subfigure[WN18] {
		\includegraphics[width=37mm,height=30mm]{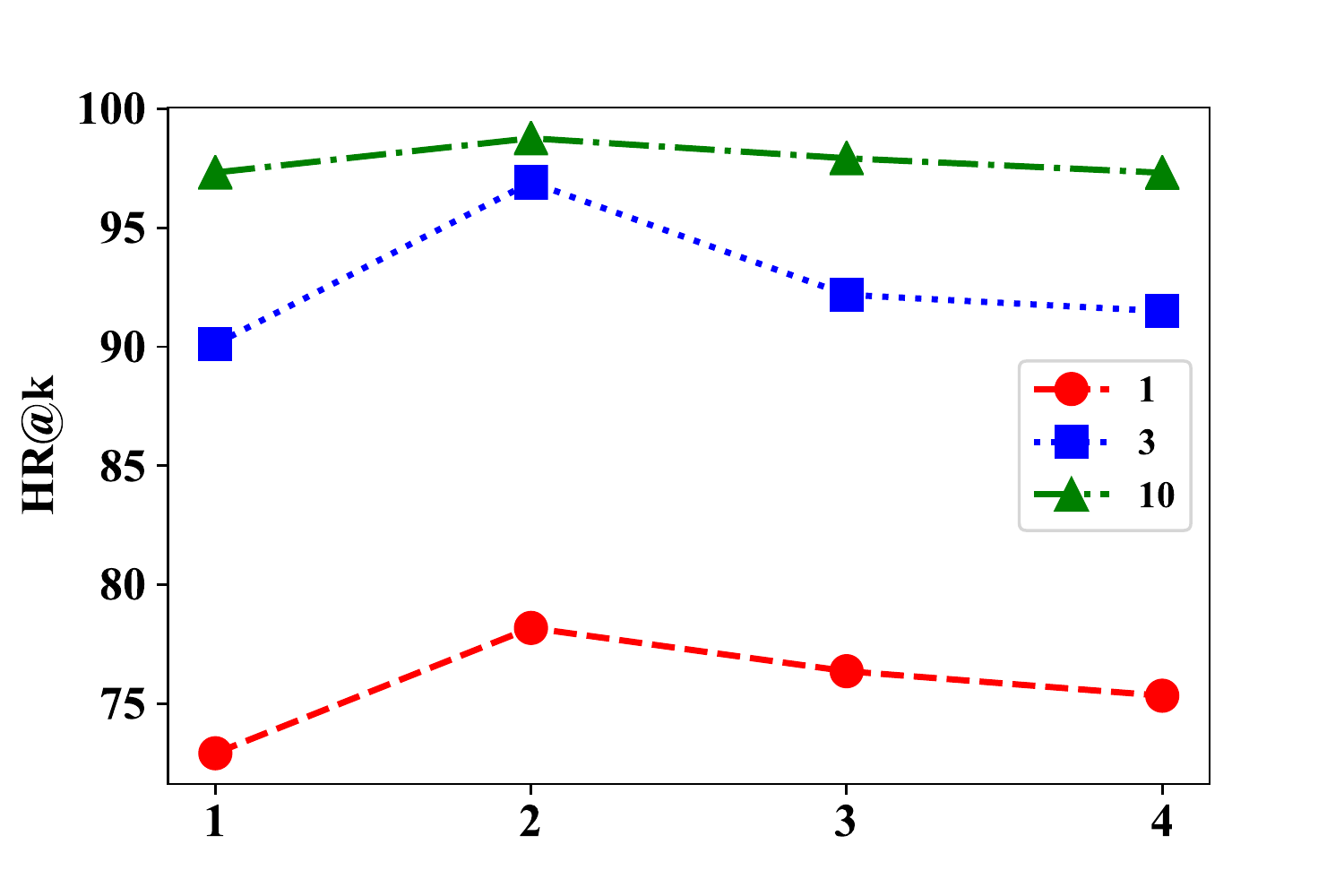}
	}
	\subfigure[FB15K] {
		\includegraphics[width=37mm,height=30mm]{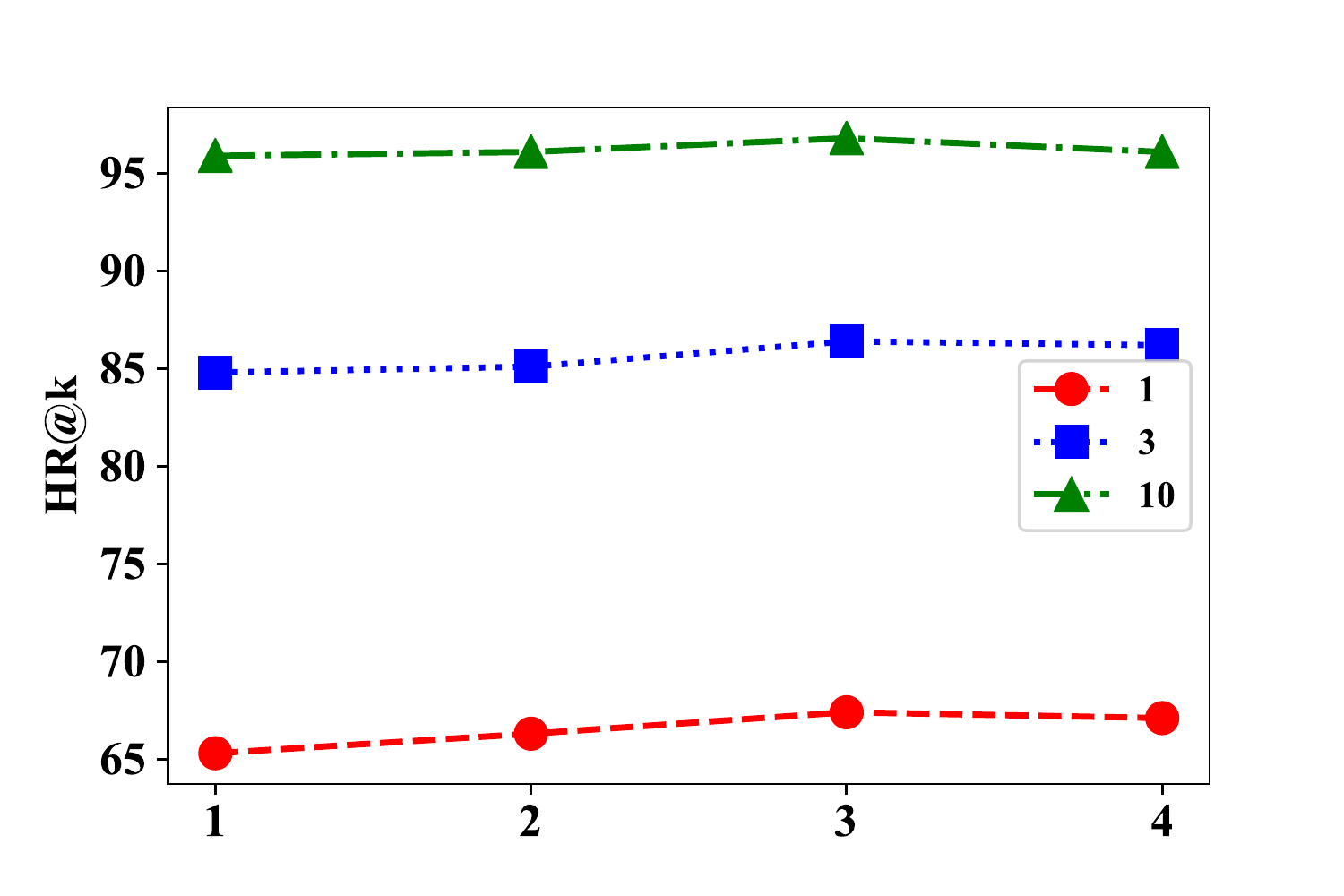}
	}
	\caption{Performance study \wrt the number of hops}
	\label{fig_layer}
\end{figure}

\begin{figure}[h]
	\centering
	\subfigure[WN18] {
		\includegraphics[width=37mm,height=30mm]{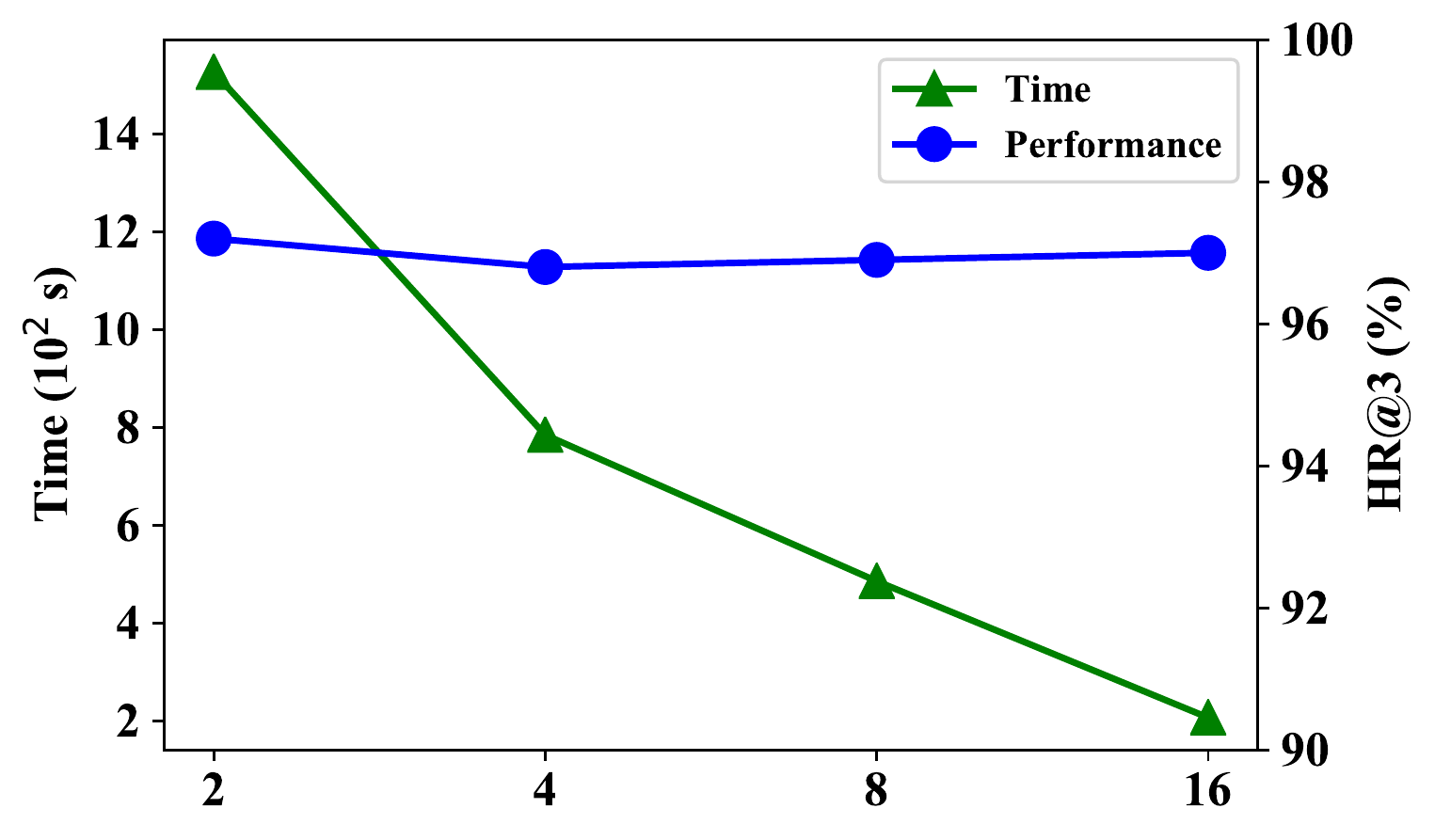}
	}
	\subfigure[FB15K] {
		\includegraphics[width=37mm,height=30mm]{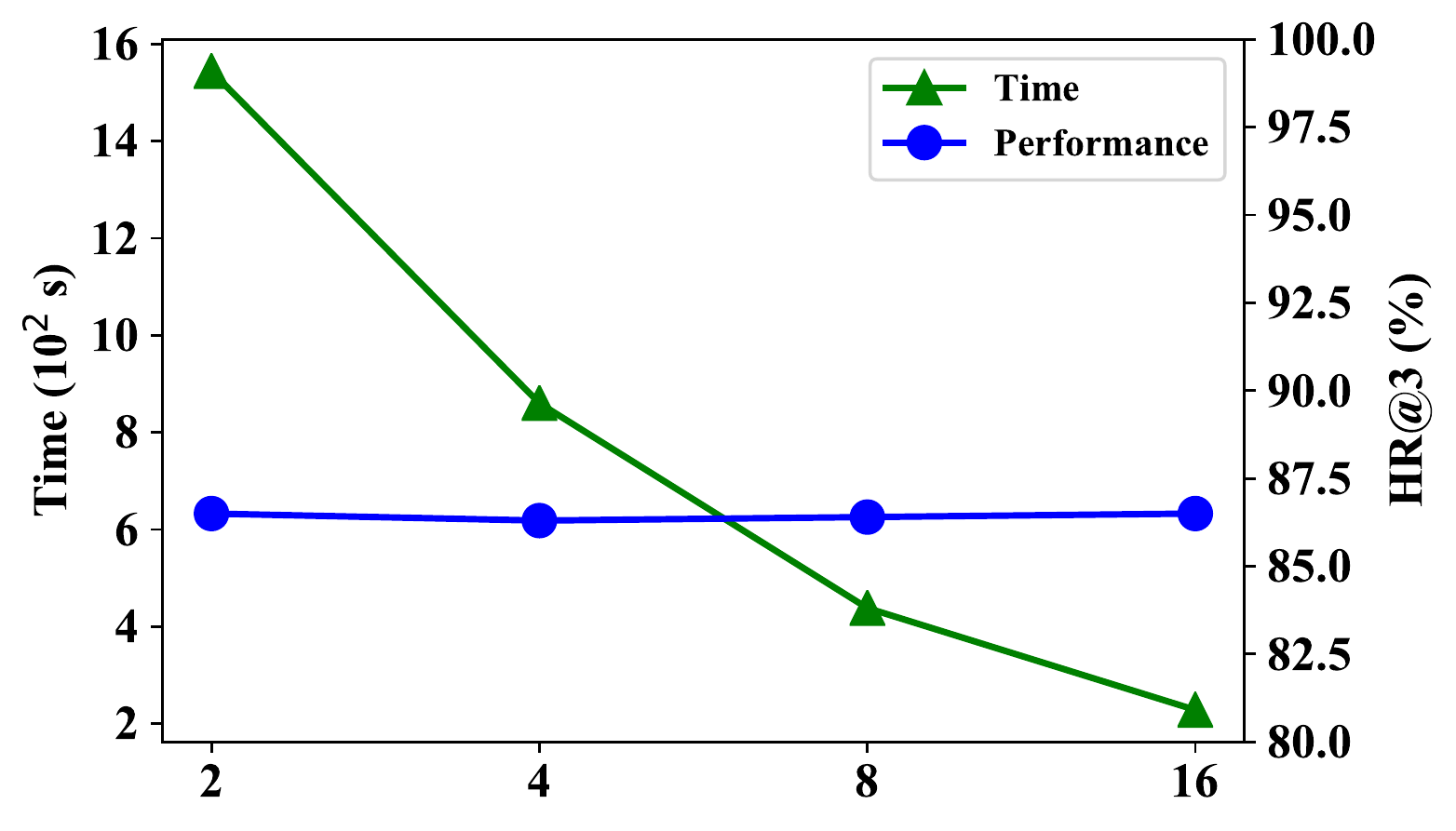}
	}
	\caption{Scalability study \wrt the number of workers}
	\label{fig_sca}
\end{figure}



\textbf{Effect of the number of hops.}
We analyze the effect of the number of hops on the link prediction task through varying it among \{1, 2, 3, 4\}. As shown in Fig.~\ref{fig_layer} , the proposed KGNN achieve the optimal performance when \# hop = 2 on WN18 and \# hop = 3 on FB15K. The results indicates high-order structure information exactly help our model learn more powerful representations, while excessive hops of neighbors would harm the performance due to the over-smoothing problem~\cite{chen2019measuring}.

\textbf{Scalability study.}
To verify the scalability of our proposed distributed KGNN framework, we report the updating time per training epoch \wrt the number of workers in Fig.~\ref{fig_sca}. As shown, the speed up in training KGNN on WN18 and FB15K is consistent as we increase the number of workers from 2 to 16. Meanwhile, it also shows that there is almost no loss of predictive performance as the number of workers increases.

\section{Conclusion}

In this paper, we proposed a novel distributed framework called KGNN for graph neural knowledge representation with GNN based encoder and knowledge aware decoder, which  jointly exploit high-order structure and attribute information together for powerful knowledge representation as well as preserve relation patterns in KGs. Furthermore, an attention mechanism is introduced to  emphasize important information for fine-grained modeling.
We implement the proposed KGNN on the distributed learning system and extensive experiments demonstrates its effectiveness and scalability.

\clearpage

\bibliography{reference}
\bibliographystyle{icml2020}





\end{document}